\pdfoutput=1

\documentclass[11pt]{article}

\usepackage{emnlp2022}

\usepackage{times}
\usepackage{latexsym}

\usepackage[T1]{fontenc}

\usepackage[utf8]{inputenc}

\usepackage{microtype}

\usepackage{amsmath}
\usepackage{amssymb}
\usepackage{booktabs}
\usepackage[capitalize,noabbrev]{cleveref}
\usepackage{fontawesome5}
\usepackage{graphicx}
\usepackage{pdfcomment}
\usepackage{subcaption}
\usepackage[utf8]{inputenc}
\usepackage[T1]{fontenc}
\usepackage{tikz}
\usepackage{CJKutf8} 
\usepackage{multirow}

\newcommand*\circled[1]{\tikz[baseline=-.3em]{\node[shape=circle,draw,inner sep=1.3pt] (char) {\scriptsize #1};}}
\newcommand{\nlpnorth}{\hspace{.2em}\faIcon{compass}}
\newcommand{\cis}{\resizebox{!}{.75em}{\faIcon{mountain}}}

\usetikzlibrary{shadings}
\pgfdeclarehorizontalshading{rainbow-left}{100bp}{%
  rgb(0bp)=(1,0,0);
  rgb(52bp)=(1,.5,0);
  rgb(66bp)=(1,1,.1);
  rgb(80bp)=(1,1,1);
  rgb(94bp)=(1,1,1);
  rgb(100bp)=(1,1,1)}
\pgfdeclarehorizontalshading{rainbow-right}{100bp}{%
  rgb(0bp)=(1,1,1);
  rgb(20bp)=(1,1,1);
  rgb(52bp)=(.1,1,1);
  rgb(66bp)=(0,.1,1);
  rgb(80bp)=(1,0,1);
  rgb(94bp)=(.5,0,.5);
  rgb(100bp)=(.5,0,.5)}

\newcommand{\RainbowLeft}{\vspace*{-1ex}\begin{tikzpicture}
\pgfsetfillopacity{0.5}
\path[shading=rainbow-left] (0,0) rectangle ++ (5cm, 1pt);
\end{tikzpicture}}

\newcommand{\RainbowRight}{\vspace*{-1ex}\begin{tikzpicture}
\pgfsetfillopacity{0.5}
\path[shading=rainbow-right] (0,0) rectangle ++ (5cm, 1pt);
\end{tikzpicture}}

\title{Spectral Probing}
\title{
Spectral Probing\\
{\vspace{-.9em}
\RainbowLeft{}
\hspace{4cm}
\RainbowRight{}
}
}

\author{
    Max M{\"u}ller-Eberstein\textsuperscript{\nlpnorth{}}  \and
    Rob van der Goot\textsuperscript{\nlpnorth{}} \and
    Barbara Plank\textsuperscript{\nlpnorth{}\cis{}\faRobot} \\
    \textsuperscript{\nlpnorth{}} Department of Computer Science, IT University of Copenhagen, Denmark \\
    \textsuperscript{\cis{}} Center for Information and Language Processing (CIS), LMU Munich, Germany \\
    \textsuperscript{\faRobot} Munich Center for Machine Learning (MCML), Munich, Germany \\
    \texttt{mamy@itu.dk, robv@itu.dk, b.plank@lmu.de} \\}

\begin{document}

\maketitle

\begin{abstract}
Linguistic information is encoded at varying timescales (subwords, phrases, etc.) and communicative levels, such as syntax and semantics. Contextualized embeddings have analogously been found to capture these phenomena at distinctive layers and frequencies. Leveraging these findings, we develop a fully learnable frequency filter to identify \textit{spectral profiles} for any given task. It enables vastly more granular analyses than prior handcrafted filters, and improves on efficiency. After demonstrating the informativeness of spectral probing over manual filters in a monolingual setting, we investigate its multilingual characteristics across seven diverse NLP tasks in six languages. Our analyses identify distinctive spectral profiles which quantify cross-task similarity in a linguistically intuitive manner, while remaining consistent across languages---highlighting their potential as robust, lightweight task descriptors.
\end{abstract}

%
%
\section{Introduction}

Analyzing the contextualized embedding representations of pre-trained language models (LMs) using lightweight probes \citep{hewitt-liang-2019-designing,voita-titov-2020-information} has identified latent features in the untuned encoders which are highly relevant to downstream NLP tasks at various layer depths \citep{tenney-etal-2019-bert}. Orthogonally, linguistic phenomena are also encoded at different timescales: i.e., rapidly changing (sub-)word-level information versus slower changing sentence or paragraph-level information. Decomposing contextualized embeddings into frequencies with different rates of change has yielded initial insights into the timescales at which these task-specific latent phenomena occur \citep{tamkin2020prism}. These findings currently rely on handcrafted spectral filters and are limited to English. To enable more efficient analyses of finer-grained, continuous frequency spectra in contextualized representations covering more tasks and languages, this work contributes:

\begin{itemize}
    \item A fully differentiable spectral probing framework for \textit{learning} which frequencies are relevant for specific NLP tasks (\cref{sec:method}).\footnote{Code at \href{https://github.com/mainlp/spectral-probing}{https://github.com/mainlp/spectral-probing}.}
    \item A multilingual probing study examining timescale characteristics of seven diverse NLP tasks across six languages (\cref{sec:experiments}).
    \item An analysis of the relationships between the spectral profiles of different tasks and their consistency across languages (\cref{sec:analysis}).
\end{itemize}

\begin{figure}
    \centering
    \pdftooltip{\includegraphics[width=.44\textwidth]{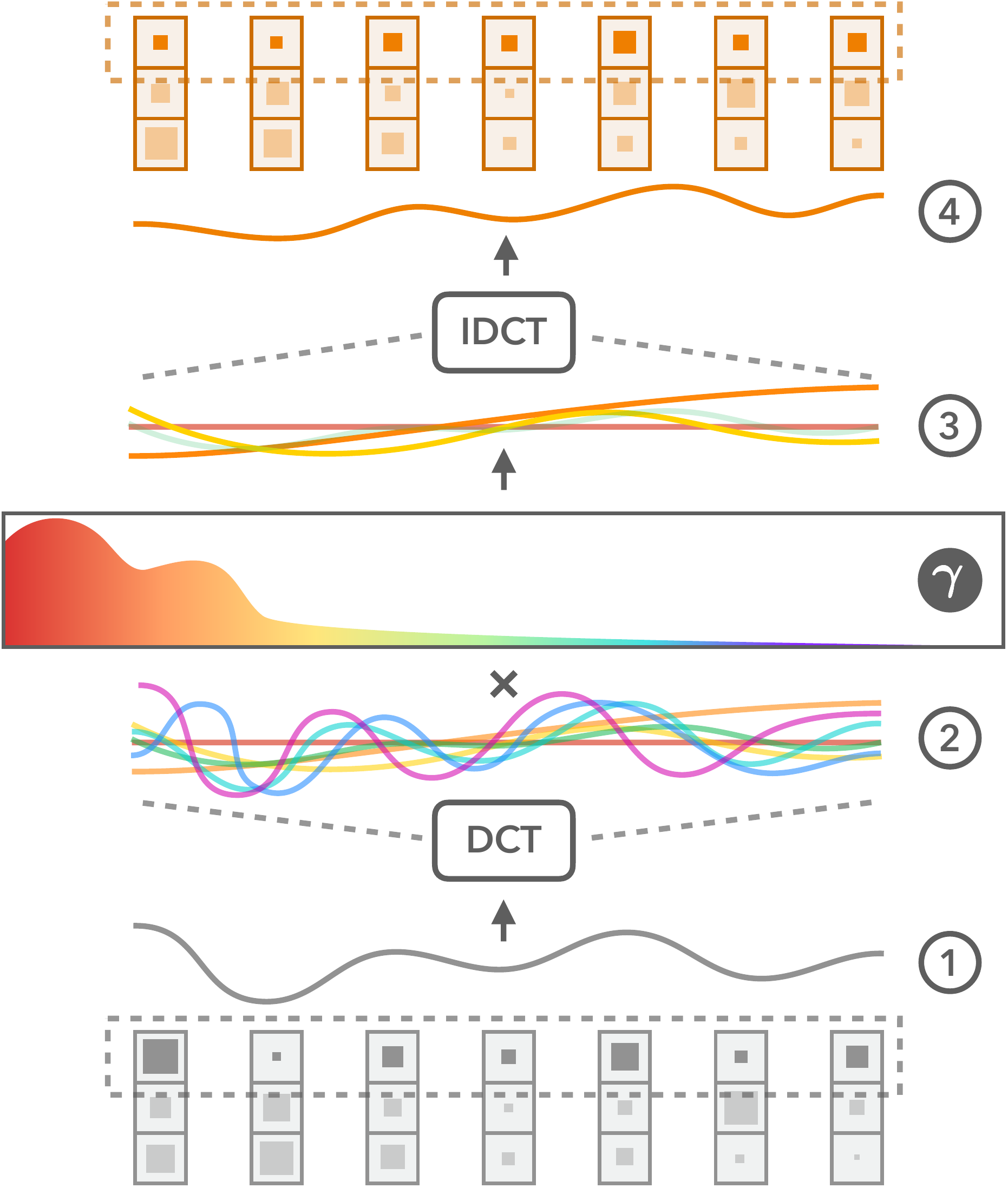}}{Screenreader Caption: Step-wise schematic of spectral probing. Step 1: A single row from a sequence of embeddings is represented as a wave built from each cell's value (low to high). Step 2: Decompose the wave into its composite frequencies (slow to fast) using DCT. Step 3: Apply the spectral probe gamma to continuously weight each frequency. Step 4: Recompose the filtered frequencies into the output sequence of embedding values using IDCT.}
    \caption{\textbf{Visualization of Spectral Probing}. Given a sequence of embedding values \circled{1}, decompose into composite frequency waves using DCT \circled{2}, apply the learned filter \circled{$\boldsymbol{\gamma}$}, retaining a subset of waves \circled{3}, for which IDCT returns the filtered sequence of values \circled{4}.}
    \label{fig:spectral-probing}
\end{figure}

%
%
\section{Probing for Spectral Profiles}\label{sec:method}

Spectral Probing (\cref{fig:spectral-probing}) builds on established signal processing methods \citep{ahmed1974dct} and recent findings on the manual frequency filtering of contextual embeddings \citep{tamkin2020prism}. The method automatically learns spectral profiles which measure the relevance of specific frequencies to a given task by amplifying or reducing contextual information with different rates of change.

\paragraph{Discrete Cosine Transform} (\citealp{ahmed1974dct}; DCT) is an invertible method for decomposing any sequence of real values $\{x_0, \dots, x_{N-1}\}$ (e.g., all values of an embedding dimension) into a weighted sum over cosine waves with different frequencies. The number of frequencies equals the sequence length $N$, as the lowest frequency wave is a constant ($k=0$) and the highest frequency wave completes one cycle every timestep ($k=N-1$). The coefficient $X_n^{(k)}$ for a wave at DCT index $k$ at timestep $n$ is calculated as:

\begin{equation}\label{eq:dct}
    \pdftooltip{X_n^{(k)} = \sum_{n=0}^{N-1} x_n \cos \left[ \frac{\pi}{N} \left(n + \frac{1}{2}\right) k \right] \hspace{.1em}\text{.}}{Screenreader Caption: X superscript k subscript n equals sum from n equals 0 to N minus 1 over x subscript n times cosine of k times pi over N times n plus one half.}
\end{equation}

Inverting the DCT (IDCT) using all $X_n^{(k)}$ will return the original sequence. However, weighting coefficients for some $k$ by $0$ will return a filtered version. Zeroing out all $k$ above a threshold will only retain lower frequencies and make values oscillate with a slow rate of change. Vice-versa, zeroing out all $k$ below a threshold will only retain higher frequencies---amplifying short-term changes.

\paragraph{Fixed-band Filters} Applying frequency filters to a sequence of contextualized embeddings extracts linguistic information at different timescales. Within this formulation, the values across each embedding dimension are gathered into a real-valued sequence to which transformations such as the DCT can be applied. In seminal work, \citet{tamkin2020prism} apply manually defined low ($k \in [0,1]$), mid-low ($k \in [2,8]$), mid ($k \in [9,33]$), mid-high ($k \in [34,129]$) and high frequency filters ($k \in [130,511]$) to English BERT embeddings \citep{devlin-etal-2019-bert} to investigate how accurately a linear probe can extract task-specific information within certain spectra. Capturing the full picture using manual, fixed-band filters is however not computationally feasible: Relevant frequencies might not lie in a contiguous band, and furthermore, frequencies can not only be turned on or off (i.e., weighted 0 or 1), but can actually be weighted continuously in $[0,1]$.

\paragraph{Learnable Filters} To capture the complete picture, we propose spectral probing which \textit{learns} a continuous weighting of frequencies relevant to a task. In effect, the spectral filter is a vector $\boldsymbol{\gamma} \in \mathbb{R}^M$ for which each entry corresponds to the weight assigned to a particular frequency. Before inverting the DCT, each $X^{(k)}_n$ is multiplied by the sigmoid-scaled weight $\gamma^{(k)} \in [0,1]$ which will then retain or filter out frequencies at index $k$. As $M$ depends on the sequence length $N$, which changes across inputs, the spectral probe dynamically scales $\boldsymbol{\gamma}$ to the length at hand using adaptive mean pooling. In practice, we set $M$ to the maximum input length for our given encoder (e.g., 512 for BERT) and shrink $\boldsymbol{\gamma}$ appropriately, as a wave cannot cycle more often than there are values. It would however be equally possible to set $M$ smaller than $N$ and interpolate the filter up to the length required. Overall, $\boldsymbol{\gamma}$ is a lightweight parameter which can be easily incorporated between the frozen encoder and probing head, and uses the existing training objective to jointly learn which frequencies to amplify or filter out.

%
%
\section{Experiments}\label{sec:experiments}


\subsection{Monolingual}\label{sec:exp-monolingual}

\paragraph{Setup} Initially, we compare spectral probing to previous fixed-band filters by reproducing the highest and lowest frequency experiments by \citet{tamkin2020prism}. These are the tasks of tagging parts-of-speech (\textsc{PoS}) in the Penn Treebank (\citealp{marcus-etal-1993-building}; PTB) as well as classifying \textsc{Topics} in the 20 Newsgroups corpus (\citealp{lang95news}; 20News).

On the modeling side, we follow \citet{tamkin2020prism} and train a linear probe \citep{alain2017linear} on top of the frozen LM encoder to classify each manually/automatically filtered contextual embedding in an input sequence. This corresponds to probing and evaluating for the amount of task-relevant information in each sub-word across a sequence (e.g., underlying topic contextualization). The bands for the five manual filters follow the original definitions (see \cref{sec:method}), and we compare them to unfiltered (\textsc{Orig}) as well as automatically filtered (\textsc{Auto}) embeddings from our spectral probe (details in \cref{app:exp-setup}).

\begin{figure}
    \centering
    \begin{subfigure}[t]{.48\textwidth}
        \pdftooltip{\includegraphics[width=.49\textwidth]{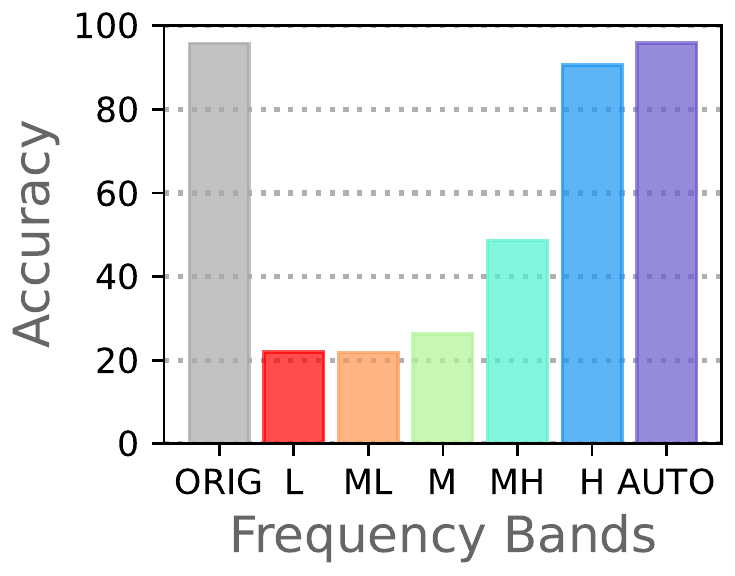}}{Screenreader Caption: PoS-tagging accuracy of each filtering strategy. Original (unfiltered): 95.8, Low: 21.9, Mid-low: 21.8, Mid: 26.2, Mid-high: 48.6, High: 90.6, Automatically filtered: 95.9.}
        \pdftooltip{\includegraphics[width=.49\textwidth]{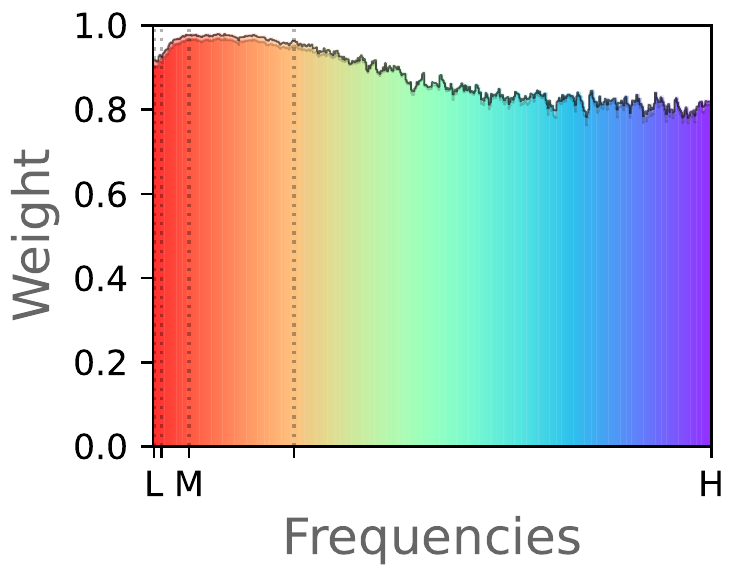}}{Screenreader Caption: Monolingual PoS-tagging spectral profile weightings. Low: 0.9-0.95, Mid: 0.9-0.95, High: 0.8-0.95.}
        \vspace{-.5em}
        \caption{PTB (\textsc{PoS})}
        \vspace{.5em}
        \label{fig:results-ptb}
    \end{subfigure}
    \begin{subfigure}[t]{.48\textwidth}
        \pdftooltip{\includegraphics[width=.49\textwidth]{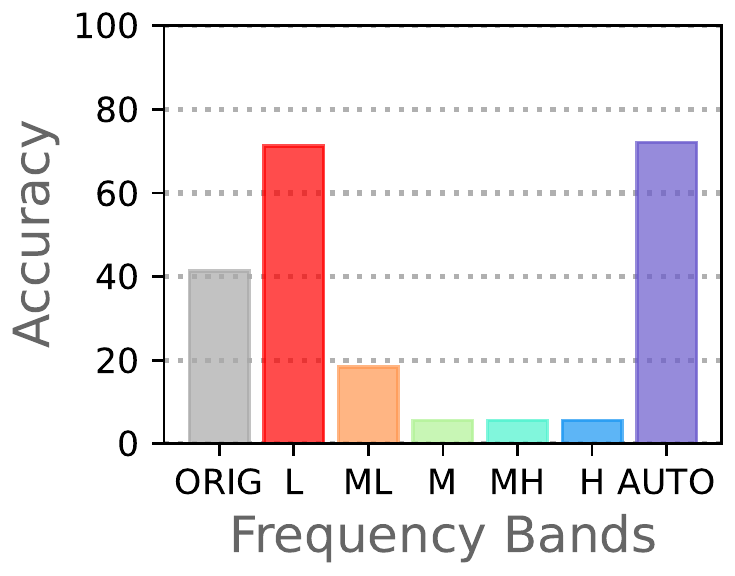}}{Screenreader Caption: Topic classification accuracy of each filtering strategy. Original (unfiltered): 41.3, Low: 71.2, Mid-low: 18.4, Mid: 5.6, Mid-high: 5.6, High: 5.6, Automatically filtered: 72.1.}
        \pdftooltip{\includegraphics[width=.49\textwidth]{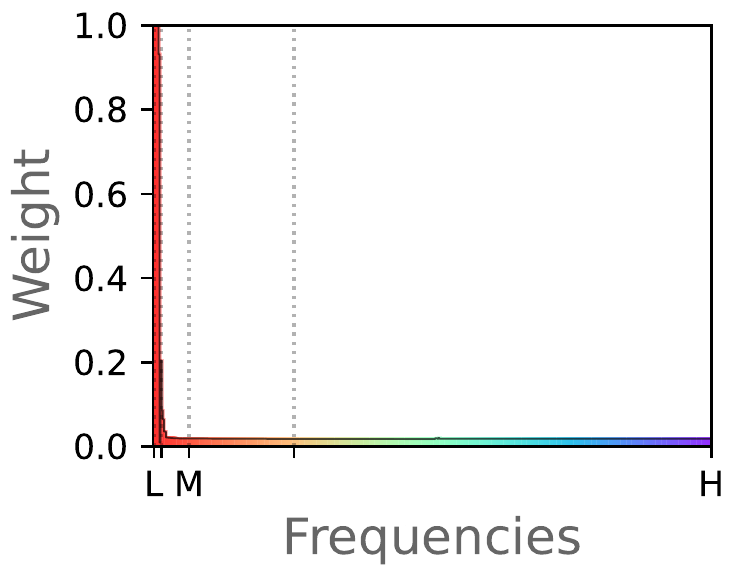}}{Screenreader Caption: Monolingual Topic classification spectral profile weightings. Low: 1.0, Mid: 0-0.1, High: 0.}
        \vspace{-.5em}
        \caption{20News (\textsc{Topic})}
        \label{fig:results-20news}
    \end{subfigure}
    \caption{\textbf{Monolingual Results on PTB and 20News.} \textsc{Acc} of unfiltered (\textsc{Orig}), low (L), mid-low (ML), mid (M), mid-high (MH), high (H), and the spectral probe's automatic filters (\textsc{Auto}) with frequency weightings.}
    \vspace{-.5em}
    \label{fig:results-monolingual}
\end{figure}

\paragraph{Results} \cref{fig:results-monolingual} shows the accuracy (\textsc{Acc}) of the six prior filtering strategies in addition to the learned frequency weightings of the spectral probe. The unfiltered and manually filtered embeddings corroborate previous findings \cite{tamkin2020prism}, with high frequencies performing best on \textsc{PoS}, and the lowest frequencies performing best on \textsc{Topic}.

The spectral probe achieves 95.9\% \textsc{Acc} for \textsc{PoS}, outperforming \textsc{Orig} by a 0.1\% margin and the best manual filter by 5.2\%. The spectral profile in \cref{fig:results-ptb} (right) sheds light on why this may be the case: While it also prioritizes high (sub-)word-level frequencies, the learned filter additionally includes surprising amounts of mid-high and lower frequencies, emphasizing the need for both local and global context to achieve high performance.

For \textsc{Topic}, the spectral probe achieves 72.1\% \textsc{Acc}, outperforming both \textsc{Orig} (41.3\%) and the fixed low-band filter (71.2\%). The learned filter (see \cref{fig:results-20news}, right) mirrors the fixed-band results: Only the lowest bands are active, while all higher ones are not. As mid-low frequencies still appear to contain weaker amounts of topic information, the soft inclusion of this region by the spectral probe could account for its performance boost. Overall, spectral probing confirms and refines frequency ranges from prior work while surfacing more detail and requiring no manual probe engineering, with only a single probing run instead of five.

\begin{table}[h]
\centering
\resizebox{.3\textwidth}{!}{
\begin{tabular}{lrr}
\toprule
\textsc{Task} & \textsc{Orig} & \textsc{Auto} \\
\midrule
\textsc{PoS} & 92.4{\footnotesize$\pm$1.9} & 92.5{\footnotesize$\pm$1.8} \\
\textsc{Dep} & 78.6{\footnotesize$\pm$4.3} & 79.3{\footnotesize$\pm$4.3} \\
NER & 88.0{\footnotesize$\pm$2.7} & 88.1{\footnotesize$\pm$2.6} \\
QA & 62.9{\footnotesize$\pm$1.6} & 67.1{\footnotesize$\pm$1.2} \\
\textsc{Senti} & 57.4{\footnotesize$\pm$0.9} & 64.3{\footnotesize$\pm$1.1}\\
\textsc{Topic} & 27.1{\footnotesize$\pm$8.1} & 37.2{\footnotesize$\pm$8.2}\\
NLI & 44.1{\footnotesize$\pm$4.1} & 56.3{\footnotesize$\pm$5.6}\\
\bottomrule
\end{tabular}
}
\caption{\textbf{Multilingual Results} (\textsc{Acc}) of unfiltered (\textsc{Orig}) and automatically filtered (\textsc{Auto}) embeddings. Means $\pm$ standard deviations over languages and random initializations (details in \cref{app:detaied-results}).}
\label{tab:results-multilingual}
\end{table}

\begin{figure}[h]
    \centering
    \hspace{-1em}
    \pdftooltip{\includegraphics[width=.45\textwidth]{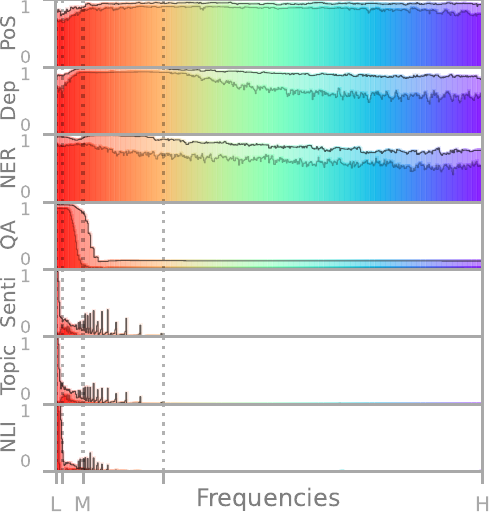}}{Screenreader Caption: Multilingual spectral profile weightings. PoS: Low: 0.8-0.9, Mid: 0.9-1.0, High: 0.9-1.0. Dep: Low: 0.7-0.8, Mid: 0.9-1.0, High: 0.5-0.8. NER: Low: 0.8-1.0, Mid: 0.7-1.0, High: 0.5-0.9. QA: Low: 0.9-1.0, Mid: 0.1-0.9, High: 0-0.2. Senti: Low: 1.0, Mid: 0.1-0.3, High: 0. Topic: Low: 1.0, Mid: 0.1-0.3, High: 0. NLI Low: 1.0, Mid: 0.1-0.3, High: 0.}
    \caption{\textbf{Spectral Profiles} of all tasks (weight per frequency), with lower and upper bounds across languages.}
    \vspace{-1em}
    \label{fig:filters}
\end{figure}

\subsection{Multilingual}\label{sec:exp-multilingual}

Leveraging spectral probing, we extend timescale analyses beyond English and investigate spectral profiles across more diverse tasks and languages.
 
\paragraph{Setup} Each experiment covers German (DE), English (EN), Spanish (ES), French (FR), Japanese (JA) and Chinese (ZH). The tasks are \textsc{PoS}-tagging and dependency relation classification (\textsc{Dep}) from Universal Dependencies \citep{ud29}; named entity recognition (NER) from WikiANN \citep{pan-etal-2017-cross}; question answering (QA) from MKQA \citep{longpre-etal-2021-mkqa}; sentiment analysis (\textsc{Senti}) and \textsc{Topic} classification from Multilingual Amazon Reviews \citep{keung-etal-2020-multilingual}; natural language inference (NLI) from XNLI \citep{conneau-etal-2018-xnli} and JSNLI \citep{yoshikoshi2020jsnli} for JA (details and examples in \cref{app:data-setup}).

For each language-task combination we train a linear probe on the unfiltered embeddings of multilingual BERT (\citealp{devlin-etal-2019-bert}; mBERT) and on the automatically filtered representations from our spectral probe. The remaining settings are identical to the monolingual setup (details in \cref{app:exp-setup}).

\paragraph{Results} \cref{tab:results-multilingual} shows equivalent or higher \textsc{Acc} for the spectral filter compared to the unfiltered embeddings for all tasks and languages. This increase is less pronounced for token-level tasks, but much larger for tasks where sequence-level information is critical. \cref{fig:filters} visualizes how \textsc{PoS}, \textsc{Dep} and \textsc{NER} retain large parts of the original spectrum, while \textsc{QA}, \textsc{Senti}, \textsc{Topic} and \textsc{NLI} appear to benefit from filtering out higher frequencies. This shows how tasks exhibit structures at different timescales and that spectral probing is able to identify these  communicative levels consistently not only in English, but also across languages---an effect which we analyze more extensively next.

%
%
\section{Spectral Profiling Analysis}\label{sec:analysis}

\begin{figure}
    \centering
    \begin{subfigure}[b]{.23\textwidth}
        \pdftooltip{\includegraphics[width=\textwidth]{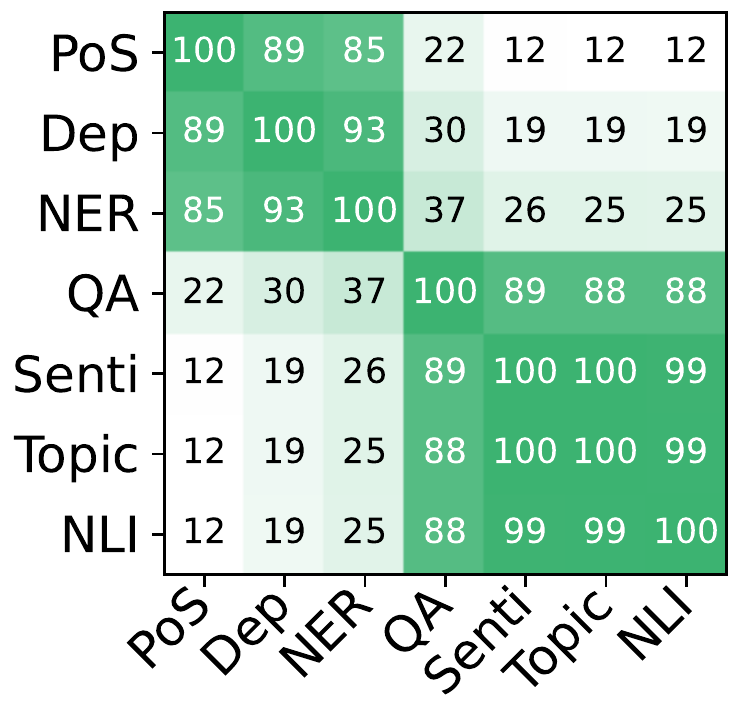}}{Screenreader Caption: Confusion matrix of task spectral profile overlap percentage. PoS/Dep: 89, PoS/NER: 85, PoS/QA: 22, PoS/Senti: 12, PoS/Topic: 12, PoS/NLI: 12, Dep/NER: 93, Dep/QA: 30, Dep/Senti: 19, Dep/Topic: 19, Dep/NLI: 19, NER/QA: 37, NER/Senti: 26, NER/Topic: 25, NER/NLI: 25, QA/Senti: 89, QA/Topic: 88, QA/NLI: 88, Senti/Topic: 100, Senti/NLI: 99, Topic/NLI: 99.}
        \caption{Tasks}
        \label{fig:distances-tasks}
    \end{subfigure}
    \begin{subfigure}[b]{.23\textwidth}
        \pdftooltip{\includegraphics[width=.9\textwidth]{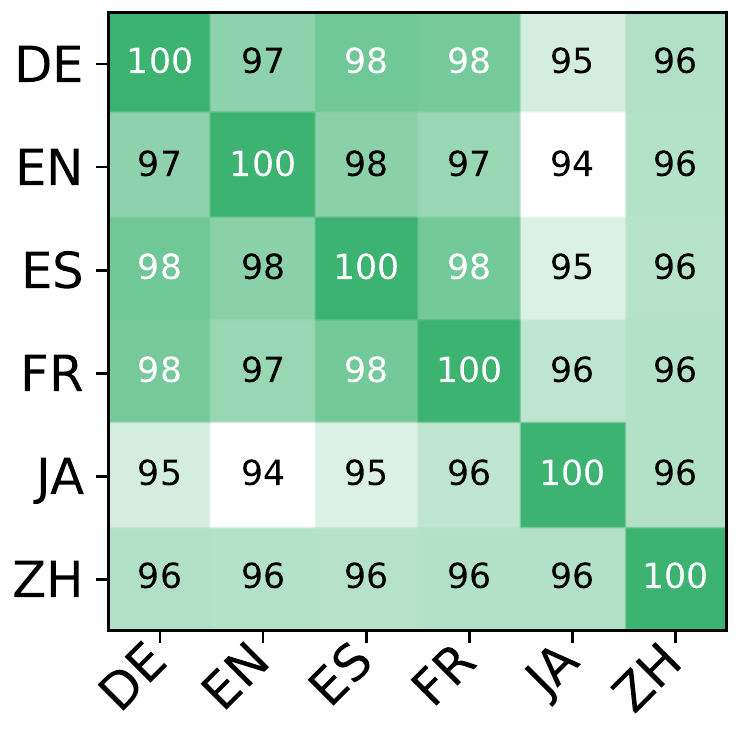}}{Screenreader Caption: Confusion matrix of cross-lingual spectral profile overlap percentage. DE/EN: 97, DE/ES: 98, DE/FR: 98, DE/JA: 95, DE/ZH: 96, EN/ES: 98, EN/FR: 97, EN/JA: 94, EN/ZH: 96, ES/FR: 98, ES/JA: 95, ES/ZH: 96, FR/JA: 96, FR/ZH: 96, JA/ZH: 96.}
        \vspace{.7em}
        \caption{Languages}
        \label{fig:distances-languages}
    \end{subfigure}
    \caption{\textbf{Filter Overlap across Tasks/Languages} as measured in percentage-normalized L1 distance.}
    \label{fig:distances}
\end{figure}

Each task's distinct spectral profile (\cref{fig:filters}) allows us to analyze their relation to the timescale hierarchy of linguistic structures, and quantify cross-task similarities within and across languages. For this we use the percentage-normalized L1 distance  (i.e., 0\%--100\% overlap) between filters (\cref{fig:distances}).

\paragraph{Cross-task Overlap} 
Overall, we observe a dichotomy between broad-frequency, token-level tasks and low-frequency, sequence-level tasks (Figures \ref{fig:filters} and \ref{fig:distances-tasks}). In addition, there appears to be a hierarchy which depends on the timescales of the linguistic structures involved. Notably, compared to prior fixed-band filters, none of the learned filters fully excludes low frequencies. For instance, high-frequency information is most important to retrieve \textsc{PoS}, but reaching the performance of the original embeddings also requires some lower-frequency information---most likely to disambiguate difficult cases based on sentence-level context.

\textsc{Dep} appears to benefit the least from both lower and higher-frequency information. Instead, the strong weight on mid-high frequencies matches the fact that dependency relations span multiple words and benefit from information at the phrase-level. \textsc{NER} sees a further decrease in high-frequency information, coupled with an uptick in lower frequencies. We hypothesize that phrase and sentence-level information become more important for disambiguating certain entity types (e.g., \texttt{ORG} and \texttt{LOC}). Across the token-level tasks this shift from higher to lower frequencies is also reflected in filter overlap which decreases from syntactic to semantic token-level tasks, while their overlap with sentence-level tasks increases (\cref{fig:distances-tasks}).

The sequence-level tasks share low-frequency spectral profiles which overlap more with each other than do the token-level tasks. In fact, \textsc{Senti} and \textsc{Topic} overlap almost perfectly (although the latter involves less mid-range frequencies). This similarity is unlikely to be the result of the shared underlying dataset as both tasks also overlap with the unrelated XNLI and JSNLI datasets. At the same time, the \textsc{PoS} and \textsc{Dep} tasks, which also share datasets, have a lower overlap despite being based on the exact same inputs. Overall, \textsc{Senti}, \textsc{Topic} and NLI all appear to rely on information which is consistent across a sequence---explaining why simple methods such as mean-pooled sentence embeddings can be effective in these scenarios.

QA provides an intermediate case: While it is reliant on low frequencies it also includes more mid-low and a small amount of higher frequency information. This is reflected in \cref{fig:distances-tasks}, where it shares more overlap with the token-level tasks than all other sequence-level tasks. Since probing for the correctness of a question-answer pair is dependent on finer-grained information than the general sentiment, topic or semantic coherence of a sequence, this inclusion of higher frequency information matches linguistic intuitions.

\paragraph{Cross-lingual Consistency} Finally, we investigate the similarity of learned spectral profiles across languages. While \cref{fig:filters} shows that there is some variance between the filters of different languages within a task, \cref{fig:distances-languages} shows that actual quantitative overlap between languages is high, ranging from 94\%--98\%. This holds even across distinctive pairs such as JA-EN which differ substantially in factors such as sub-word length and distance between syntactic dependents. This strong consistency highlights the potential for spectral profiles to provide language-agnostic features for task characterization and comparison.

%
%
\section{Conclusion}

Linguistic information at different timescales is an, as of yet, underexplored dimension in contextualized embeddings. We propose a fully differentiable \textit{spectral probe} which automatically learns to weigh frequencies that are relevant to a specific task and improves over prior fixed-band filters by capturing continuous mixtures over frequencies (\cref{sec:method}). This enables us to not only outperform the manual filters while using one probe instead of five, but to also identify that high-frequency tasks still benefit from low-frequency information (\cref{sec:exp-monolingual}).

Extending spectral probing to seven tasks in six languages, we trained task-specific filters which outperformed the original, unfiltered embeddings. The resulting spectral profiles furthermore shed light onto how linguistic information at different timescales relates to different task types (\cref{sec:exp-multilingual}). They not only match the linguistic intuitions underlying each task, but also enable quantitative comparisons between them. The analysis of the filters' overlap surfaced a clear dichotomy between token and sequence-level tasks, but also highlighted intersecting frequency ranges which contain information relevant across task types. Finally, the language-agnostic nature of these spectral profiles highlights future avenues towards more robust task descriptors (\cref{sec:analysis}).

%
%
\section*{Limitations}\label{sec:limitations}

Our experiments cover a diverse, but non-exhaustive set of NLP tasks and languages. While more extensive than prior related work \citep{tenney-etal-2019-bert,tamkin2020prism}, we elaborate in the following regarding the motivation of the final setup: As the aim of our study was to investigate the cross-lingual properties of the underexplored timescale dimension of contextualized representations, the set of languages and tasks used in our experiments emphasizes consistency across languages. This limits us to high-resource languages for which datasets covering every task are available. However, with cross-lingual stability confirmed in our experiments, the study of lower-resourced languages is a clear avenue for future research.

Despite using a set of well-established datasets, it is important to keep data quality in mind when interpreting the results---even for these high-resource languages. In our initial exploratory data analyses, we identified and confirmed limitations known to the original dataset authors in that many include silver, or weakly filtered annotations driven by automatic matching and translation (e.g., WikiANN, XNLI, JSNLI). As we are less interested in benchmarking performance and rather focus on the feasibility and analysis of our spectral profiles, individual data instances of lesser quality should however be of limited concern. \cref{app:data-setup} details how each dataset was constructed originally, and also how it was pre-processed by us, such that results can be interpreted in the appropriate context.

In terms of modeling, we hope that future work will investigate spectral probes and their resulting task profiles across more encoder models with different architectures and pre-training strategies. Finally, while we have demonstrated spectral profiles to be suitable for characterizing different tasks consistently across languages, future research could supplement them with other descriptors such as embedding layer depth in order to identify even more distinctive profiles.

\section*{Ethics Statement}

Given the theoretical nature and wide applicability of this work---both in terms of data domains and model architectures---it is difficult to anticipate broader impacts and future ethical implications. In general, benefits and harms in the field of probing originate from the information being investigated: While we are interested in linguistic timescale characteristics, probe-like methods have also been applied to protected attributes of data subjects in order to, for example, de-bias LMs \citep{ravfogel-etal-2020-null}. Since this process involves personal information, any experiments extracting such characteristics should be sufficiently vetted for ethical acceptability. With spectral profiles being a relatively broad descriptor however, we do not expect them to identify frequencies exclusive to personal information or to replace existing, domain-specific probing methods.

\section*{Acknowledgements}
Thanks to \nlpnorth{} NLPnorth and \cis{} MaiNLP Lab for their insightful feedback, particularly Elisa Bassignana, Joris Baan, Mike Zhang, as well as to ITU's High-performance Computing Team. Additional thanks to the anonymous reviewers for their helpful comments. MME and BP are supported by the Independent Research Fund Denmark (DFF) Sapere Aude grant 9063-00077B. BP is supported by the ERC Consolidator Grant DIALECT 101043235.

\bibliography{anthology,references}

\begin{thebibliography}{29}
\expandafter\ifx\csname natexlab\endcsname\relax\def\natexlab#1{#1}\fi

\bibitem[{Ahmed et~al.(1974)Ahmed, Natarajan, and Rao}]{ahmed1974dct}
Nasir Ahmed, T.~Natarajan, and Kamisetty~Ramamohan Rao. 1974.
\newblock \href {https://doi.org/10.1109/T-C.1974.223784} {Discrete cosine
  transform}.
\newblock \emph{IEEE Transactions on Computers}, C-23(1):90--93.

\bibitem[{Alain and Bengio(2017)}]{alain2017linear}
Guillaume Alain and Yoshua Bengio. 2017.
\newblock \href {https://openreview.net/forum?id=HJ4-rAVtl} {Understanding
  intermediate layers using linear classifier probes}.
\newblock In \emph{5th International Conference on Learning Representations,
  {ICLR} 2017, Toulon, France, April 24-26, 2017, Workshop Track Proceedings}.
  OpenReview.net.

\bibitem[{Asahara et~al.(2018)Asahara, Kanayama, Tanaka, Miyao, Uematsu, Mori,
  Matsumoto, Omura, and Murawaki}]{asahara-etal-2018-universal}
Masayuki Asahara, Hiroshi Kanayama, Takaaki Tanaka, Yusuke Miyao, Sumire
  Uematsu, Shinsuke Mori, Yuji Matsumoto, Mai Omura, and Yugo Murawaki. 2018.
\newblock \href {https://aclanthology.org/L18-1287} {{U}niversal {D}ependencies
  version 2 for {J}apanese}.
\newblock In \emph{Proceedings of the Eleventh International Conference on
  Language Resources and Evaluation ({LREC} 2018)}, Miyazaki, Japan. European
  Language Resources Association (ELRA).

\bibitem[{Bowman et~al.(2015)Bowman, Angeli, Potts, and
  Manning}]{bowman-etal-2015-large}
Samuel~R. Bowman, Gabor Angeli, Christopher Potts, and Christopher~D. Manning.
  2015.
\newblock \href {https://doi.org/10.18653/v1/D15-1075} {A large annotated
  corpus for learning natural language inference}.
\newblock In \emph{Proceedings of the 2015 Conference on Empirical Methods in
  Natural Language Processing}, pages 632--642, Lisbon, Portugal. Association
  for Computational Linguistics.

\bibitem[{Brants et~al.(2004)Brants, Dipper, Eisenberg, Hansen-Schirra,
  K{\"o}nig, Lezius, Rohrer, Smith, and Uszkoreit}]{brants2004tiger}
Sabine Brants, Stefanie Dipper, Peter Eisenberg, Silvia Hansen-Schirra, Esther
  K{\"o}nig, Wolfgang Lezius, Christian Rohrer, George Smith, and Hans
  Uszkoreit. 2004.
\newblock {TIGER}: Linguistic interpretation of a german corpus.
\newblock \emph{Research on language and computation}, 2(4):597--620.

\bibitem[{Conneau et~al.(2018)Conneau, Rinott, Lample, Williams, Bowman,
  Schwenk, and Stoyanov}]{conneau-etal-2018-xnli}
Alexis Conneau, Ruty Rinott, Guillaume Lample, Adina Williams, Samuel Bowman,
  Holger Schwenk, and Veselin Stoyanov. 2018.
\newblock \href {https://doi.org/10.18653/v1/D18-1269} {{XNLI}: Evaluating
  cross-lingual sentence representations}.
\newblock In \emph{Proceedings of the 2018 Conference on Empirical Methods in
  Natural Language Processing}, pages 2475--2485, Brussels, Belgium.
  Association for Computational Linguistics.

\bibitem[{Devlin et~al.(2019)Devlin, Chang, Lee, and
  Toutanova}]{devlin-etal-2019-bert}
Jacob Devlin, Ming-Wei Chang, Kenton Lee, and Kristina Toutanova. 2019.
\newblock \href {https://doi.org/10.18653/v1/N19-1423} {{BERT}: Pre-training of
  deep bidirectional transformers for language understanding}.
\newblock In \emph{Proceedings of the 2019 Conference of the North {A}merican
  Chapter of the Association for Computational Linguistics: Human Language
  Technologies, Volume 1 (Long and Short Papers)}, pages 4171--4186,
  Minneapolis, Minnesota. Association for Computational Linguistics.

\bibitem[{Guillaume et~al.(2019)Guillaume, de~Marneffe, and
  Perrier}]{guillaume2019conversion}
Bruno Guillaume, Marie-Catherine de~Marneffe, and Guy Perrier. 2019.
\newblock Conversion et am{\'e}liorations de corpus du {F}ran{\c{c}}ais
  annot{\'e}s en {U}niversal {D}ependencies.
\newblock \emph{Traitement Automatique des Langues}, 60(2):71--95.

\bibitem[{Harris et~al.(2020)Harris, Millman, van~der Walt, Gommers, Virtanen,
  Cournapeau, Wieser, Taylor, Berg, Smith, Kern, Picus, Hoyer, van Kerkwijk,
  Brett, Haldane, del R{\'{i}}o, Wiebe, Peterson, G{\'{e}}rard-Marchant,
  Sheppard, Reddy, Weckesser, Abbasi, Gohlke, and Oliphant}]{numpy}
Charles~R. Harris, K.~Jarrod Millman, St{\'{e}}fan~J. van~der Walt, Ralf
  Gommers, Pauli Virtanen, David Cournapeau, Eric Wieser, Julian Taylor,
  Sebastian Berg, Nathaniel~J. Smith, Robert Kern, Matti Picus, Stephan Hoyer,
  Marten~H. van Kerkwijk, Matthew Brett, Allan Haldane, Jaime~Fern{\'{a}}ndez
  del R{\'{i}}o, Mark Wiebe, Pearu Peterson, Pierre G{\'{e}}rard-Marchant,
  Kevin Sheppard, Tyler Reddy, Warren Weckesser, Hameer Abbasi, Christoph
  Gohlke, and Travis~E. Oliphant. 2020.
\newblock \href {https://doi.org/10.1038/s41586-020-2649-2} {Array programming
  with {NumPy}}.
\newblock \emph{Nature}, 585(7825):357--362.

\bibitem[{Hewitt and Liang(2019)}]{hewitt-liang-2019-designing}
John Hewitt and Percy Liang. 2019.
\newblock \href {https://doi.org/10.18653/v1/D19-1275} {Designing and
  interpreting probes with control tasks}.
\newblock In \emph{Proceedings of the 2019 Conference on Empirical Methods in
  Natural Language Processing and the 9th International Joint Conference on
  Natural Language Processing (EMNLP-IJCNLP)}, pages 2733--2743, Hong Kong,
  China. Association for Computational Linguistics.

\bibitem[{Hu(2018)}]{ziyang2018torchdct}
Ziyang Hu. 2018.
\newblock {D}iscrete {C}osine {T}ransform for {P}y{T}orch.
\newblock \url{https://github.com/zh217/torch-dct}.

\bibitem[{Hunter(2007)}]{matplotlib}
J.~D. Hunter. 2007.
\newblock \href {https://doi.org/10.1109/MCSE.2007.55} {Matplotlib: A 2d
  graphics environment}.
\newblock \emph{Computing in Science \& Engineering}, 9(3):90--95.

\bibitem[{Keung et~al.(2020)Keung, Lu, Szarvas, and
  Smith}]{keung-etal-2020-multilingual}
Phillip Keung, Yichao Lu, Gy{\"o}rgy Szarvas, and Noah~A. Smith. 2020.
\newblock \href {https://doi.org/10.18653/v1/2020.emnlp-main.369} {The
  multilingual {A}mazon reviews corpus}.
\newblock In \emph{Proceedings of the 2020 Conference on Empirical Methods in
  Natural Language Processing (EMNLP)}, pages 4563--4568, Online. Association
  for Computational Linguistics.

\bibitem[{Kingma and Ba(2014)}]{kingma2014adam}
Diederik~P. Kingma and Jimmy Ba. 2014.
\newblock \href {https://arxiv.org/abs/1412.6980} {Adam: A method for
  stochastic optimization}.
\newblock \emph{Computing Research Repository}, arxiv:1412.6980.
\newblock Version 9.

\bibitem[{Lang(1995)}]{lang95news}
Ken Lang. 1995.
\newblock Newsweeder: Learning to filter netnews.
\newblock In \emph{Proceedings of the Twelfth International Conference on
  Machine Learning}, pages 331--339.
\newblock Retrieved from \url{http://qwone.com/~jason/20Newsgroups/} on Feb
  10th, 2022.

\bibitem[{Longpre et~al.(2021)Longpre, Lu, and Daiber}]{longpre-etal-2021-mkqa}
Shayne Longpre, Yi~Lu, and Joachim Daiber. 2021.
\newblock \href {https://doi.org/10.1162/tacl_a_00433} {{MKQA}: A
  linguistically diverse benchmark for multilingual open domain question
  answering}.
\newblock \emph{Transactions of the Association for Computational Linguistics},
  9:1389--1406.

\bibitem[{Marcus et~al.(1993)Marcus, Santorini, and
  Marcinkiewicz}]{marcus-etal-1993-building}
Mitchell~P. Marcus, Beatrice Santorini, and Mary~Ann Marcinkiewicz. 1993.
\newblock \href {https://aclanthology.org/J93-2004} {Building a large annotated
  corpus of {E}nglish: The {P}enn {T}reebank}.
\newblock \emph{Computational Linguistics}, 19(2):313--330.

\bibitem[{McDonald et~al.(2013)McDonald, Nivre, Quirmbach-Brundage, Goldberg,
  Das, Ganchev, Hall, Petrov, Zhang, T{\"a}ckstr{\"o}m, Bedini,
  Bertomeu~Castell{\'o}, and Lee}]{mcdonald-etal-2013-universal}
Ryan McDonald, Joakim Nivre, Yvonne Quirmbach-Brundage, Yoav Goldberg, Dipanjan
  Das, Kuzman Ganchev, Keith Hall, Slav Petrov, Hao Zhang, Oscar
  T{\"a}ckstr{\"o}m, Claudia Bedini, N{\'u}ria Bertomeu~Castell{\'o}, and
  Jungmee Lee. 2013.
\newblock \href {https://aclanthology.org/P13-2017} {{U}niversal {D}ependency
  annotation for multilingual parsing}.
\newblock In \emph{Proceedings of the 51st Annual Meeting of the Association
  for Computational Linguistics (Volume 2: Short Papers)}, pages 92--97, Sofia,
  Bulgaria. Association for Computational Linguistics.

\bibitem[{Pan et~al.(2017)Pan, Zhang, May, Nothman, Knight, and
  Ji}]{pan-etal-2017-cross}
Xiaoman Pan, Boliang Zhang, Jonathan May, Joel Nothman, Kevin Knight, and Heng
  Ji. 2017.
\newblock \href {https://doi.org/10.18653/v1/P17-1178} {Cross-lingual name
  tagging and linking for 282 languages}.
\newblock In \emph{Proceedings of the 55th Annual Meeting of the Association
  for Computational Linguistics (Volume 1: Long Papers)}, pages 1946--1958,
  Vancouver, Canada. Association for Computational Linguistics.

\bibitem[{Paszke et~al.(2019)Paszke, Gross, Massa, Lerer, Bradbury, Chanan,
  Killeen, Lin, Gimelshein, Antiga, Desmaison, Kopf, Yang, DeVito, Raison,
  Tejani, Chilamkurthy, Steiner, Fang, Bai, and Chintala}]{pytorch}
Adam Paszke, Sam Gross, Francisco Massa, Adam Lerer, James Bradbury, Gregory
  Chanan, Trevor Killeen, Zeming Lin, Natalia Gimelshein, Luca Antiga, Alban
  Desmaison, Andreas Kopf, Edward Yang, Zachary DeVito, Martin Raison, Alykhan
  Tejani, Sasank Chilamkurthy, Benoit Steiner, Lu~Fang, Junjie Bai, and Soumith
  Chintala. 2019.
\newblock \href
  {http://papers.neurips.cc/paper/9015-pytorch-an-imperative-style-high-performance-deep-learning-library.pdf}
  {Py{T}orch: An imperative style, high-performance deep learning library}.
\newblock In \emph{Advances in Neural Information Processing Systems 32}, pages
  8024--8035. Curran Associates, Inc.

\bibitem[{Ravfogel et~al.(2020)Ravfogel, Elazar, Gonen, Twiton, and
  Goldberg}]{ravfogel-etal-2020-null}
Shauli Ravfogel, Yanai Elazar, Hila Gonen, Michael Twiton, and Yoav Goldberg.
  2020.
\newblock \href {https://doi.org/10.18653/v1/2020.acl-main.647} {Null it out:
  Guarding protected attributes by iterative nullspace projection}.
\newblock In \emph{Proceedings of the 58th Annual Meeting of the Association
  for Computational Linguistics}, pages 7237--7256, Online. Association for
  Computational Linguistics.

\bibitem[{Shen et~al.(2016)Shen, McDonald, Zeman, and Qi}]{ud_chinese_gsd_2016}
Mo~Shen, Ryan McDonald, Daniel Zeman, and Peng Qi. 2016.
\newblock {UD\_Chinese-GSD}.
\newblock \url{https://github.com/UniversalDependencies/UD_Chinese-GSD}.

\bibitem[{Silveira et~al.(2014)Silveira, Dozat, de~Marneffe, Bowman, Connor,
  Bauer, and Manning}]{silveira-etal-2014-gold}
Natalia Silveira, Timothy Dozat, Marie-Catherine de~Marneffe, Samuel Bowman,
  Miriam Connor, John Bauer, and Chris Manning. 2014.
\newblock \href
  {http://www.lrec-conf.org/proceedings/lrec2014/pdf/1089_Paper.pdf} {A gold
  standard dependency corpus for {E}nglish}.
\newblock In \emph{Proceedings of the Ninth International Conference on
  Language Resources and Evaluation ({LREC}'14)}, pages 2897--2904, Reykjavik,
  Iceland. European Language Resources Association (ELRA).

\bibitem[{Tamkin et~al.(2020)Tamkin, Jurafsky, and Goodman}]{tamkin2020prism}
Alex Tamkin, Dan Jurafsky, and Noah Goodman. 2020.
\newblock \href
  {https://proceedings.neurips.cc/paper/2020/file/3acb2a202ae4bea8840224e6fce16fd0-Paper.pdf}
  {Language through a prism: A spectral approach for multiscale language
  representations}.
\newblock In \emph{Advances in Neural Information Processing Systems},
  volume~33, pages 5492--5504. Curran Associates, Inc.

\bibitem[{Tenney et~al.(2019)Tenney, Das, and Pavlick}]{tenney-etal-2019-bert}
Ian Tenney, Dipanjan Das, and Ellie Pavlick. 2019.
\newblock \href {https://doi.org/10.18653/v1/P19-1452} {{BERT} rediscovers the
  classical {NLP} pipeline}.
\newblock In \emph{Proceedings of the 57th Annual Meeting of the Association
  for Computational Linguistics}, pages 4593--4601, Florence, Italy.
  Association for Computational Linguistics.

\bibitem[{Voita and Titov(2020)}]{voita-titov-2020-information}
Elena Voita and Ivan Titov. 2020.
\newblock \href {https://doi.org/10.18653/v1/2020.emnlp-main.14}
  {Information-theoretic probing with minimum description length}.
\newblock In \emph{Proceedings of the 2020 Conference on Empirical Methods in
  Natural Language Processing (EMNLP)}, pages 183--196, Online. Association for
  Computational Linguistics.

\bibitem[{Wolf et~al.(2020)Wolf, Debut, Sanh, Chaumond, Delangue, Moi, Cistac,
  Rault, Louf, Funtowicz, Davison, Shleifer, von Platen, Ma, Jernite, Plu, Xu,
  Le~Scao, Gugger, Drame, Lhoest, and Rush}]{wolf-etal-2020-transformers}
Thomas Wolf, Lysandre Debut, Victor Sanh, Julien Chaumond, Clement Delangue,
  Anthony Moi, Pierric Cistac, Tim Rault, Remi Louf, Morgan Funtowicz, Joe
  Davison, Sam Shleifer, Patrick von Platen, Clara Ma, Yacine Jernite, Julien
  Plu, Canwen Xu, Teven Le~Scao, Sylvain Gugger, Mariama Drame, Quentin Lhoest,
  and Alexander Rush. 2020.
\newblock \href {https://doi.org/10.18653/v1/2020.emnlp-demos.6} {Transformers:
  State-of-the-art natural language processing}.
\newblock In \emph{Proceedings of the 2020 Conference on Empirical Methods in
  Natural Language Processing: System Demonstrations}, pages 38--45, Online.
  Association for Computational Linguistics.

\bibitem[{Yoshikoshi et~al.(2020)Yoshikoshi, Kawahara, and
  Kurohashi}]{yoshikoshi2020jsnli}
Takumi Yoshikoshi, Daisuke Kawahara, and Sadao Kurohashi. 2020.
\newblock Multilingualization of a natural language inference dataset using
  machine translation.
\newblock \emph{Proceedings of the 244th Meeting of Natural Language
  Processing}, pages 1--8.
\newblock ({T}ranslated from Japanese original).

\bibitem[{Zeman et~al.(2021)Zeman, Nivre, Abrams, Ackermann, Aepli, Aghaei,
  Agi{\'c}, Ahmadi, Ahrenberg, Ajede, Aleksandravi{\v c}i{\=u}t{\.e}, Alfina,
  Antonsen, Aplonova, Aquino, Aragon, Aranzabe, Ar{\i}can, Arnard{\'o}ttir,
  Arutie, Arwidarasti, Asahara, Aslan, Ateyah, Atmaca, Attia, Atutxa,
  Augustinus, Badmaeva, Balasubramani, Ballesteros, Banerjee, Bank,
  Barbu~Mititelu, Barkarson, Basile, Basmov, Batchelor, Bauer, Bedir,
  Bengoetxea, Berk, Berzak, Bhat, Bhat, Biagetti, Bick, Bielinskien{\.e},
  Bjarnad{\'o}ttir, Blokland, Bobicev, Boizou, Borges~V{\"o}lker, B{\"o}rstell,
  Bosco, Bouma, Bowman, Boyd, Braggaar, Brokait{\.e}, Burchardt, Candito,
  Caron, Caron, Cassidy, Cavalcanti, Cebiro{\u g}lu~Eryi{\u g}it, Cecchini,
  Celano, {\v C}{\'e}pl{\"o}, Cesur, Cetin, {\c C}etino{\u g}lu, Chalub,
  Chauhan, Chi, Chika, Cho, Choi, Chun, Chung, Cignarella, Cinkov{\'a},
  Collomb, {\c C}{\"o}ltekin, Connor, Courtin, Cristescu, Daniel, Davidson,
  de~Marneffe, de~Paiva, Derin, de~Souza, Diaz~de Ilarraza, Dickerson,
  Dinakaramani, Di~Nuovo, Dione, Dirix, Dobrovoljc, Dozat, Droganova, Dwivedi,
  Eckhoff, Eiche, Eli, Elkahky, Ephrem, Erina, Erjavec, Etienne, Evelyn,
  Facundes, Farkas, Ferdaousi, Fernanda, Fernandez~Alcalde, Foster, Freitas,
  Fujita, Gajdo{\v s}ov{\'a}, Galbraith, Garcia, G{\"a}rdenfors, Garza,
  Gerardi, Gerdes, Ginter, Godoy, Goenaga, Gojenola, G{\"o}k{\i}rmak, Goldberg,
  G{\'o}mez~Guinovart, Gonz{\'a}lez~Saavedra, Grici{\=u}t{\.e}, Grioni, Grobol,
  Gr{\= u}z{\={\i}}tis, Guillaume, Guillot-Barbance, G{\"u}ng{\"o}r, Habash,
  Hafsteinsson, Haji{\v c}, Haji{\v c}~jr., H{\"a}m{\"a}l{\"a}inen,
  H{\`a}~M{\~y}, Han, Hanifmuti, Hardwick, Harris, Haug, Heinecke, Hellwig,
  Hennig, Hladk{\'a}, Hlav{\'a}{\v c}ov{\'a}, Hociung, Hohle, Huber, Hwang,
  Ikeda, Ingason, Ion, Irimia, Ishola, Ito, Jannat, Jel{\'{\i}}nek, Jha,
  Johannsen, J{\'o}nsd{\'o}ttir, J{\o}rgensen, Juutinen, K, Ka{\c s}{\i}kara,
  Kaasen, Kabaeva, Kahane, Kanayama, Kanerva, Kara, Katz, Kayadelen, Kenney,
  Kettnerov{\'a}, Kirchner, Klementieva, Klyachko, K{\"o}hn, K{\"o}ksal,
  Kopacewicz, Korkiakangas, K{\"o}se, Kotsyba, Kovalevskait{\.e}, Krek,
  Krishnamurthy, K{\"u}bler, Kuyruk{\c c}u, Kuzgun, Kwak, Laippala, Lam,
  Lambertino, Lando, Larasati, Lavrentiev, Lee, L{\^e}~H{\`{\^o}}ng, Lenci,
  Lertpradit, Leung, Levina, Li, Li, Li, Li, Lim, Lima~Padovani, Lind{\'e}n,
  Ljube{\v s}i{\'c}, Loginova, Lusito, Luthfi, Luukko, Lyashevskaya, Lynn,
  Macketanz, Mahamdi, Maillard, Makazhanov, Mandl, Manning, Manurung, Mar{\c
  s}an, M{\u a}r{\u a}nduc, Mare{\v c}ek, Marheinecke, Mart{\'{\i}}nez~Alonso,
  Mart{\'{\i}}n-Rodr{\'{\i}}guez, Martins, Ma{\v s}ek, Matsuda, Matsumoto,
  Mazzei, {McDonald}, {McGuinness}, Mendon{\c c}a, Merzhevich, Miekka,
  Mischenkova, Misirpashayeva, Missil{\"a}, Mititelu, Mitrofan, Miyao,
  Mojiri~Foroushani, Moln{\'a}r, Moloodi, Montemagni, More, Moreno~Romero,
  Moretti, Mori, Mori, Morioka, Moro, Mortensen, Moskalevskyi, Muischnek,
  Munro, Murawaki, M{\"u}{\"u}risep, Nainwani, Nakhl{\'e},
  Navarro~Hor{\~n}iacek, Nedoluzhko, Ne{\v s}pore-B{\=e}rzkalne, Nevaci,
  Nguy{\~{\^e}}n~Th{\d i}, Nguy{\~{\^e}}n Th{\d i}~Minh, Nikaido, Nikolaev,
  Nitisaroj, Nourian, Nurmi, Ojala, Ojha, Ol{\'u}{\`o}kun, Omura, Onwuegbuzia,
  Osenova, {\"O}stling, {\O}vrelid, {\"O}zate{\c s}, {\"O}z{\c c}elik,
  {\"O}zg{\"u}r, {\"O}zt{\"u}rk~Ba{\c s}aran, Park, Partanen, Pascual,
  Passarotti, Patejuk, Paulino-Passos, Peljak-{\L}api{\'n}ska, Peng, Perez,
  Perkova, Perrier, Petrov, Petrova, Phelan, Piitulainen, Pirinen, Pitler,
  Plank, Poibeau, Ponomareva, Popel, Pretkalni{\c n}a, Pr{\'e}vost, Prokopidis,
  Przepi{\'o}rkowski, Puolakainen, Pyysalo, Qi, R{\"a}{\"a}bis, Rademaker,
  Rahoman, Rama, Ramasamy, Ramisch, Rashel, Rasooli, Ravishankar, Real, Rebeja,
  Reddy, Regnault, Rehm, Riabov, Rie{\ss}ler, Rimkut{\.e}, Rinaldi, Rituma,
  Rizqiyah, Rocha, R{\"o}gnvaldsson, Romanenko, Rosa, Roșca, Rovati, Rudina,
  Rueter, R{\'u}narsson, Sadde, Safari, Sagot, Sahala, Saleh, Salomoni,
  Samard{\v z}i{\'c}, Samson, Sanguinetti, San{\i}yar, S{\"a}rg,
  Saul{\={\i}}te, Sawanakunanon, Saxena, Scannell, Scarlata, Schneider,
  Schuster, Schwartz, Seddah, Seeker, Seraji, Shahzadi, Shen, Shimada, Shirasu,
  Shishkina, Shohibussirri, Sichinava, Siewert, Sigurðsson, Silveira,
  Silveira, Simi, Simionescu, Simk{\'o}, {\v S}imkov{\'a}, Simov, Skachedubova,
  Smith, Soares-Bastos, Sourov, Spadine, Sprugnoli, Steingr{\'{\i}}msson,
  Stella, Straka, Strickland, Strnadov{\'a}, Suhr, Sulestio, Sulubacak, Suzuki,
  Sz{\'a}nt{\'o}, Taguchi, Taji, Takahashi, Tamburini, Tan, Tanaka, Tanaya,
  Tella, Tellier, Testori, Thomas, Torga, Toska, Trosterud, Trukhina, Tsarfaty,
  T{\"u}rk, Tyers, Uematsu, Untilov, Ure{\v s}ov{\'a}, Uria, Uszkoreit, Utka,
  Vajjala, van~der Goot, Vanhove, van Niekerk, van Noord, Varga, Villemonte
  de~la Clergerie, Vincze, Vlasova, Wakasa, Wallenberg, Wallin, Walsh, Wang,
  Washington, Wendt, Widmer, Wijono, Williams, Wir{\'e}n, Wittern, Woldemariam,
  Wong, Wr{\'o}blewska, Yako, Yamashita, Yamazaki, Yan, Yasuoka, Yavrumyan,
  Yenice, Y{\i}ld{\i}z, Yu, Yuliawati, {\v Z}abokrtsk{\'y}, Zahra, Zeldes,
  Zhou, Zhu, Zhuravleva, and Ziane}]{ud29}
Daniel Zeman, Joakim Nivre, Mitchell Abrams, Elia Ackermann, No{\"e}mi Aepli,
  Hamid Aghaei, {\v Z}eljko Agi{\'c}, Amir Ahmadi, Lars Ahrenberg,
  Chika~Kennedy Ajede, Gabriel{\.e} Aleksandravi{\v c}i{\=u}t{\.e}, Ika Alfina,
  Lene Antonsen, Katya Aplonova, Angelina Aquino, Carolina Aragon, Maria~Jesus
  Aranzabe, Bilge~Nas Ar{\i}can, {\t H}{\'o}runn Arnard{\'o}ttir, Gashaw
  Arutie, Jessica~Naraiswari Arwidarasti, Masayuki Asahara, Deniz~Baran Aslan,
  Luma Ateyah, Furkan Atmaca, Mohammed Attia, Aitziber Atutxa, Liesbeth
  Augustinus, Elena Badmaeva, Keerthana Balasubramani, Miguel Ballesteros, Esha
  Banerjee, Sebastian Bank, Verginica Barbu~Mititelu, Starkaður Barkarson,
  Rodolfo Basile, Victoria Basmov, Colin Batchelor, John Bauer, Seyyit~Talha
  Bedir, Kepa Bengoetxea, G{\"o}zde Berk, Yevgeni Berzak, Irshad~Ahmad Bhat,
  Riyaz~Ahmad Bhat, Erica Biagetti, Eckhard Bick, Agn{\.e} Bielinskien{\.e},
  Krist{\'{\i}}n Bjarnad{\'o}ttir, Rogier Blokland, Victoria Bobicev,
  Lo{\"{\i}}c Boizou, Emanuel Borges~V{\"o}lker, Carl B{\"o}rstell, Cristina
  Bosco, Gosse Bouma, Sam Bowman, Adriane Boyd, Anouck Braggaar, Kristina
  Brokait{\.e}, Aljoscha Burchardt, Marie Candito, Bernard Caron, Gauthier
  Caron, Lauren Cassidy, Tatiana Cavalcanti, G{\"u}l{\c s}en Cebiro{\u
  g}lu~Eryi{\u g}it, Flavio~Massimiliano Cecchini, Giuseppe G.~A. Celano,
  Slavom{\'{\i}}r {\v C}{\'e}pl{\"o}, Neslihan Cesur, Savas Cetin, {\"O}zlem
  {\c C}etino{\u g}lu, Fabricio Chalub, Shweta Chauhan, Ethan Chi, Taishi
  Chika, Yongseok Cho, Jinho Choi, Jayeol Chun, Juyeon Chung, Alessandra~T.
  Cignarella, Silvie Cinkov{\'a}, Aur{\'e}lie Collomb, {\c C}a{\u g}r{\i} {\c
  C}{\"o}ltekin, Miriam Connor, Marine Courtin, Mihaela Cristescu, Philemon
  Daniel, Elizabeth Davidson, Marie-Catherine de~Marneffe, Valeria de~Paiva,
  Mehmet~Oguz Derin, Elvis de~Souza, Arantza Diaz~de Ilarraza, Carly Dickerson,
  Arawinda Dinakaramani, Elisa Di~Nuovo, Bamba Dione, Peter Dirix, Kaja
  Dobrovoljc, Timothy Dozat, Kira Droganova, Puneet Dwivedi, Hanne Eckhoff,
  Sandra Eiche, Marhaba Eli, Ali Elkahky, Binyam Ephrem, Olga Erina, Toma{\v z}
  Erjavec, Aline Etienne, Wograine Evelyn, Sidney Facundes, Rich{\'a}rd Farkas,
  Jannatul Ferdaousi, Mar{\'{\i}}lia Fernanda, Hector Fernandez~Alcalde,
  Jennifer Foster, Cl{\'a}udia Freitas, Kazunori Fujita, Katar{\'{\i}}na
  Gajdo{\v s}ov{\'a}, Daniel Galbraith, Marcos Garcia, Moa G{\"a}rdenfors,
  Sebastian Garza, Fabr{\'{\i}}cio~Ferraz Gerardi, Kim Gerdes, Filip Ginter,
  Gustavo Godoy, Iakes Goenaga, Koldo Gojenola, Memduh G{\"o}k{\i}rmak, Yoav
  Goldberg, Xavier G{\'o}mez~Guinovart, Berta Gonz{\'a}lez~Saavedra, Bernadeta
  Grici{\=u}t{\.e}, Matias Grioni, Lo{\"{\i}}c Grobol, Normunds Gr{\=
  u}z{\={\i}}tis, Bruno Guillaume, C{\'e}line Guillot-Barbance, Tunga
  G{\"u}ng{\"o}r, Nizar Habash, Hinrik Hafsteinsson, Jan Haji{\v c}, Jan
  Haji{\v c}~jr., Mika H{\"a}m{\"a}l{\"a}inen, Linh H{\`a}~M{\~y}, Na-Rae Han,
  Muhammad~Yudistira Hanifmuti, Sam Hardwick, Kim Harris, Dag Haug, Johannes
  Heinecke, Oliver Hellwig, Felix Hennig, Barbora Hladk{\'a}, Jaroslava
  Hlav{\'a}{\v c}ov{\'a}, Florinel Hociung, Petter Hohle, Eva Huber, Jena
  Hwang, Takumi Ikeda, Anton~Karl Ingason, Radu Ion, Elena Irimia, {\d
  O}l{\'a}j{\'{\i}}d{\'e} Ishola, Kaoru Ito, Siratun Jannat, Tom{\'a}{\v s}
  Jel{\'{\i}}nek, Apoorva Jha, Anders Johannsen, Hildur J{\'o}nsd{\'o}ttir,
  Fredrik J{\o}rgensen, Markus Juutinen, Sarveswaran K, H{\"u}ner Ka{\c
  s}{\i}kara, Andre Kaasen, Nadezhda Kabaeva, Sylvain Kahane, Hiroshi Kanayama,
  Jenna Kanerva, Neslihan Kara, Boris Katz, Tolga Kayadelen, Jessica Kenney,
  V{\'a}clava Kettnerov{\'a}, Jesse Kirchner, Elena Klementieva, Elena
  Klyachko, Arne K{\"o}hn, Abdullatif K{\"o}ksal, Kamil Kopacewicz, Timo
  Korkiakangas, Mehmet K{\"o}se, Natalia Kotsyba, Jolanta Kovalevskait{\.e},
  Simon Krek, Parameswari Krishnamurthy, Sandra K{\"u}bler, O{\u g}uzhan
  Kuyruk{\c c}u, Asl{\i} Kuzgun, Sookyoung Kwak, Veronika Laippala, Lucia Lam,
  Lorenzo Lambertino, Tatiana Lando, Septina~Dian Larasati, Alexei Lavrentiev,
  John Lee, Phuong L{\^e}~H{\`{\^o}}ng, Alessandro Lenci, Saran Lertpradit,
  Herman Leung, Maria Levina, Cheuk~Ying Li, Josie Li, Keying Li, Yuan Li,
  {KyungTae} Lim, Bruna Lima~Padovani, Krister Lind{\'e}n, Nikola Ljube{\v
  s}i{\'c}, Olga Loginova, Stefano Lusito, Andry Luthfi, Mikko Luukko, Olga
  Lyashevskaya, Teresa Lynn, Vivien Macketanz, Menel Mahamdi, Jean Maillard,
  Aibek Makazhanov, Michael Mandl, Christopher Manning, Ruli Manurung,
  B{\"u}{\c s}ra Mar{\c s}an, C{\u a}t{\u a}lina M{\u a}r{\u a}nduc, David
  Mare{\v c}ek, Katrin Marheinecke, H{\'e}ctor Mart{\'{\i}}nez~Alonso, Lorena
  Mart{\'{\i}}n-Rodr{\'{\i}}guez, Andr{\'e} Martins, Jan Ma{\v s}ek, Hiroshi
  Matsuda, Yuji Matsumoto, Alessandro Mazzei, Ryan {McDonald}, Sarah
  {McGuinness}, Gustavo Mendon{\c c}a, Tatiana Merzhevich, Niko Miekka, Karina
  Mischenkova, Margarita Misirpashayeva, Anna Missil{\"a}, C{\u a}t{\u a}lin
  Mititelu, Maria Mitrofan, Yusuke Miyao, {AmirHossein} Mojiri~Foroushani,
  Judit Moln{\'a}r, Amirsaeid Moloodi, Simonetta Montemagni, Amir More, Laura
  Moreno~Romero, Giovanni Moretti, Keiko~Sophie Mori, Shinsuke Mori, Tomohiko
  Morioka, Shigeki Moro, Bjartur Mortensen, Bohdan Moskalevskyi, Kadri
  Muischnek, Robert Munro, Yugo Murawaki, Kaili M{\"u}{\"u}risep, Pinkey
  Nainwani, Mariam Nakhl{\'e}, Juan~Ignacio Navarro~Hor{\~n}iacek, Anna
  Nedoluzhko, Gunta Ne{\v s}pore-B{\=e}rzkalne, Manuela Nevaci, Luong
  Nguy{\~{\^e}}n~Th{\d i}, Huy{\`{\^e}}n Nguy{\~{\^e}}n Th{\d i}~Minh,
  Yoshihiro Nikaido, Vitaly Nikolaev, Rattima Nitisaroj, Alireza Nourian, Hanna
  Nurmi, Stina Ojala, Atul~Kr. Ojha, Ad{\'e}day{\d o} Ol{\'u}{\`o}kun, Mai
  Omura, Emeka Onwuegbuzia, Petya Osenova, Robert {\"O}stling, Lilja
  {\O}vrelid, {\c S}aziye~Bet{\"u}l {\"O}zate{\c s}, Merve {\"O}z{\c c}elik,
  Arzucan {\"O}zg{\"u}r, Balk{\i}z {\"O}zt{\"u}rk~Ba{\c s}aran, Hyunji~Hayley
  Park, Niko Partanen, Elena Pascual, Marco Passarotti, Agnieszka Patejuk,
  Guilherme Paulino-Passos, Angelika Peljak-{\L}api{\'n}ska, Siyao Peng,
  Cenel-Augusto Perez, Natalia Perkova, Guy Perrier, Slav Petrov, Daria
  Petrova, Jason Phelan, Jussi Piitulainen, Tommi~A Pirinen, Emily Pitler,
  Barbara Plank, Thierry Poibeau, Larisa Ponomareva, Martin Popel, Lauma
  Pretkalni{\c n}a, Sophie Pr{\'e}vost, Prokopis Prokopidis, Adam
  Przepi{\'o}rkowski, Tiina Puolakainen, Sampo Pyysalo, Peng Qi, Andriela
  R{\"a}{\"a}bis, Alexandre Rademaker, Mizanur Rahoman, Taraka Rama, Loganathan
  Ramasamy, Carlos Ramisch, Fam Rashel, Mohammad~Sadegh Rasooli, Vinit
  Ravishankar, Livy Real, Petru Rebeja, Siva Reddy, Mathilde Regnault, Georg
  Rehm, Ivan Riabov, Michael Rie{\ss}ler, Erika Rimkut{\.e}, Larissa Rinaldi,
  Laura Rituma, Putri Rizqiyah, Luisa Rocha, Eir{\'{\i}}kur R{\"o}gnvaldsson,
  Mykhailo Romanenko, Rudolf Rosa, Valentin Roșca, Davide Rovati, Olga Rudina,
  Jack Rueter, Kristj{\'a}n R{\'u}narsson, Shoval Sadde, Pegah Safari,
  Beno{\^{\i}}t Sagot, Aleksi Sahala, Shadi Saleh, Alessio Salomoni, Tanja
  Samard{\v z}i{\'c}, Stephanie Samson, Manuela Sanguinetti, Ezgi San{\i}yar,
  Dage S{\"a}rg, Baiba Saul{\={\i}}te, Yanin Sawanakunanon, Shefali Saxena,
  Kevin Scannell, Salvatore Scarlata, Nathan Schneider, Sebastian Schuster,
  Lane Schwartz, Djam{\'e} Seddah, Wolfgang Seeker, Mojgan Seraji, Syeda
  Shahzadi, Mo~Shen, Atsuko Shimada, Hiroyuki Shirasu, Yana Shishkina, Muh
  Shohibussirri, Dmitry Sichinava, Janine Siewert, Einar~Freyr Sigurðsson,
  Aline Silveira, Natalia Silveira, Maria Simi, Radu Simionescu, Katalin
  Simk{\'o}, M{\'a}ria {\v S}imkov{\'a}, Kiril Simov, Maria Skachedubova, Aaron
  Smith, Isabela Soares-Bastos, Shafi Sourov, Carolyn Spadine, Rachele
  Sprugnoli, Stein{\t h}{\'o}r Steingr{\'{\i}}msson, Antonio Stella, Milan
  Straka, Emmett Strickland, Jana Strnadov{\'a}, Alane Suhr, Yogi~Lesmana
  Sulestio, Umut Sulubacak, Shingo Suzuki, Zsolt Sz{\'a}nt{\'o}, Chihiro
  Taguchi, Dima Taji, Yuta Takahashi, Fabio Tamburini, Mary Ann~C. Tan, Takaaki
  Tanaka, Dipta Tanaya, Samson Tella, Isabelle Tellier, Marinella Testori,
  Guillaume Thomas, Liisi Torga, Marsida Toska, Trond Trosterud, Anna Trukhina,
  Reut Tsarfaty, Utku T{\"u}rk, Francis Tyers, Sumire Uematsu, Roman Untilov,
  Zde{\v n}ka Ure{\v s}ov{\'a}, Larraitz Uria, Hans Uszkoreit, Andrius Utka,
  Sowmya Vajjala, Rob van~der Goot, Martine Vanhove, Daniel van Niekerk,
  Gertjan van Noord, Viktor Varga, Eric Villemonte de~la Clergerie, Veronika
  Vincze, Natalia Vlasova, Aya Wakasa, Joel~C. Wallenberg, Lars Wallin, Abigail
  Walsh, Jing~Xian Wang, Jonathan~North Washington, Maximilan Wendt, Paul
  Widmer, Sri~Hartati Wijono, Seyi Williams, Mats Wir{\'e}n, Christian Wittern,
  Tsegay Woldemariam, Tak-sum Wong, Alina Wr{\'o}blewska, Mary Yako, Kayo
  Yamashita, Naoki Yamazaki, Chunxiao Yan, Koichi Yasuoka, Marat~M. Yavrumyan,
  Arife~Bet{\"u}l Yenice, Olcay~Taner Y{\i}ld{\i}z, Zhuoran Yu, Arlisa
  Yuliawati, Zden{\v e}k {\v Z}abokrtsk{\'y}, Shorouq Zahra, Amir Zeldes,
  He~Zhou, Hanzhi Zhu, Anna Zhuravleva, and Rayan Ziane. 2021.
\newblock \href {http://hdl.handle.net/11234/1-4611} {Universal dependencies
  2.9}.
\newblock {LINDAT}/{CLARIAH}-{CZ} digital library at the Institute of Formal
  and Applied Linguistics ({{\'U}FAL}), Faculty of Mathematics and Physics,
  Charles University.

\end{thebibliography}
\bibliographystyle{acl_natbib}

%
%
\appendix

\section*{Appendix}

%
%
\section{Data Setup}\label{app:data-setup}

In the following, we provide details about the versions, splits and pre-processing of each dataset. Additionally, we present example instances together with their token/sequence-level annotations in \cref{tab:dataSamples} (in English, where available). In our experiments, each model is tuned on the training split and only evaluated on the validation split as we are not interested in obtaining state-of-the-art results, but rather aim to analyze overall performance patterns across tasks. We use the original splits where provided and generate our own otherwise.

\begin{table*}
\centering
\resizebox{.9\textwidth}{!}{
\begin{tabular}{l p{9.25cm} l}
    \toprule
    \multicolumn{3}{l}{\textsc{Token-level Tasks}} \\
    \midrule
    PTB & \multicolumn{2}{l}{\begin{tabular}{l l l l l l l l l} In & Tokyo & , & trading & is & halted & during &  lunchtime & . \\ \texttt{IN} & \texttt{NNP} & \texttt{,} & \texttt{NN} & \texttt{VBZ} & \texttt{VBN} & \texttt{IN} & \texttt{NN} & \texttt{.} \\ \end{tabular}}\\
    \midrule
    UD & \multicolumn{2}{l}{\begin{tabular}{l l l l l l l l l l l} Can & rabbits & and & chickens & live & together & ? \\ \texttt{AUX} & \texttt{NOUN} & \texttt{CCONJ} & \texttt{NOUN} & \texttt{VERB} & \texttt{ADV} & \texttt{PUNCT}\\ \texttt{aux} & \texttt{nsubj} & \texttt{cc} & \texttt{conj} & \texttt{root} & \texttt{advmod} & \texttt{punct} \\\end{tabular}} \\
    \midrule
    WikiANN \hspace{1em} &\multicolumn{2}{l}{ \begin{tabular}{lllllllllllllllll} The & Zeros & formed & in & Chula & Vista & in & 1976 & . \\ \texttt{B-ORG} & \texttt{I-ORG} & \texttt{O} & \texttt{O} & \texttt{B-LOC} & \texttt{I-LOC} & \texttt{O} & \texttt{O} & \texttt{O}\\ \end{tabular}} \\
    \midrule
    \multicolumn{3}{l}{\textsc{Sequence-level Tasks}} \\
    \midrule
    \multirow{2}*{MKQA} & when did love become a part of marriage? $\;$ | $\;$ 18th century & \texttt{1 (true)}\\
         & when did love become a part of marriage? $\;$ | $\;$ 2016 & \texttt{0 (false)} \\
    \midrule
    \multirow{2}*{AMR} & \multirow{2}*{All socks had large holes after a few months.} & \texttt{apparel} \\
                & & \texttt{negative} \\
    \midrule
    \multirow{2}*{20News} & [...] Does anyone know how to size cold gas roll control &  \multirow{2}*{\texttt{sci.space}} \\
    & thruster tanks for sounding rockets? [...] & \\
    \midrule
    XNLI & I've got more than a job. $\;$ | $\;$ I don't have a job or any hobby. & \texttt{contradiction} \\
    \midrule 
    \multirow{3}*{JSNLI} & \begin{CJK}{UTF8}{min}{\small地下鉄を待っている間に本を読む男。}|\hspace{.5em}{\small 男は地下にいる。}\end{CJK} &  \multirow{3}*{\texttt{entailment}} \\
    & The man reads a book while waiting for the subway. & \\
    & The man is underground. & \\
    \bottomrule
\end{tabular}}
\caption{\textbf{Example Dataset Instances} annotated with respective token/sequence-level \texttt{labels}.}
\label{tab:dataSamples}
\end{table*}

\paragraph{Penn Treebank \citep{marcus-etal-1993-building}} We use Penn Treebank version 2 (PTB) as published in OntoNotes 4.0. Sections 02-21 were used for training, section 22 for validation, and section 23 for test, totaling 30,060, 1,336 and 1,640 instances respectively. The label space covers 48 part-of-speech tags. Note that \citet{tamkin2020prism} use PTB version 3 in their experiments which we were unable to obtain due to licensing constraints. As such the exact data and splits may differ.

\paragraph{Universal Dependencies \citep{ud29}} From Universal Dependencies version 2.9 (UD), we select the following treebanks: German-GSD~\cite{brants2004tiger}, English-EWT~\cite{silveira-etal-2014-gold}, Spanish-GSD~\cite{mcdonald-etal-2013-universal}, French-GSD~\cite{guillaume2019conversion}, Japanese-GSD~\cite{asahara-etal-2018-universal}, Chinese-GSD~\cite{ud_chinese_gsd_2016} with standard splits, totaling 66,040 training and 6,683 validation instances. The label set comprises the 17 UPOS classes and the 36 dependency relations which can occur between a word and its head.

\paragraph{WikiANN \citep{pan-etal-2017-cross}} This dataset contains silver NER data for 282 languages which was extracted from Wikipedia using URL references as a proxy for named entities. It contains the entity types location (\texttt{LOC}), person (\texttt{PER}) and organization (\texttt{ORG}) which are annotated in BIO-format. Our experiments use the existing data splits with 20,000 training and 10,000 validation instances.

\paragraph{MKQA \citep{longpre-etal-2021-mkqa}} Multilingual Knowledge
Questions and Answer (MKQA) is an open-domain question answering dataset which covers 10,000 questions and their corresponding answers in an aligned corpus spanning 26 languages. After removing unanswerable questions, we use each correct QA pair to generate an additional incorrect pair for the same question, yielding a total set of 13,516 instances used in our experiments. To generate an incorrect answer, we sample an alternative answer of the same type (e.g., time, number) which does not equal the correct answer. Finally, we randomly split the data 80/20 into training and validation portions for which the instances are aligned across languages (i.e., the same questions and answers). The final task is a binary classification task for whether a QA pair is true or false, with a random baseline of 50\%.


\paragraph{Multilingual Amazon Reviews \citep{keung-etal-2020-multilingual}} MAR are used for both sentiment analysis and topic classification. For \textsc{Senti}, we convert the 1--5 star rating into $\{1, 2\} \rightarrow$ \texttt{negative}, $\{3\} \rightarrow$ \texttt{neutral} and $\{4, 5\} \rightarrow$ \texttt{positive}. For \textsc{Topic}, we consider the 30 product categories as topics. All original splits are kept, resulting in 200,000 training and 5,000 validation instances per language.

\paragraph{20 Newsgroups \citep{lang95news}} This dataset contains English emails from 20 newsgroups and their corresponding topics. We use the \texttt{bydate}-version which is sorted by date and removes duplicate entries and email headers (which give away the topic). Of the official training and testing data, we subdivide the former 11,314 instances into an 80/20 train/validation split. Note that there may differences to the version used in \citet{tamkin2020prism} due to alternative splitting strategies.

\paragraph{XNLI \citep{conneau-etal-2018-xnli}} 
The Cross-lingual Natural Language Inference (XNLI) dataset covers 15 languages translated from and including English (as it lacks Japanese data, we supplement it with JSNLI). The task is to identify the relation between a premise and a hypothesis as: \texttt{contradiction}, \texttt{entailment} or \texttt{neutral}. Our setups use the original training and validation splits with 392,702 and 2,490 input pairs respectively.

\paragraph{JSNLI \citep{yoshikoshi2020jsnli}} This dataset contains premise-hypothesis pairs from the Stanford Natural Language Inference corpus \citep{bowman-etal-2015-large} which were translated automatically into Japanese and filtered for correctness. It contains 533,005 training and 3,916 validation instances with the same three classes as XNLI.

%
%
\section{Experiment Setup}\label{app:exp-setup}

\paragraph{Models} In the monolingual English experiments, we use \texttt{bert-base-cased} (\citealp{devlin-etal-2019-bert}; BERT) following \citet{tamkin2020prism}. For the multilingual experiments we use \texttt{bert-base-multilingual-cased} (\citealp{devlin-etal-2019-bert}; mBERT). For both LMs, we use respective checkpoints from the Transformer library's model hub \citep{wolf-etal-2020-transformers}.

Manual, fixed-band filters as well as the automatically learned filters are applied to the contextualized embeddings produced by the last layer of either model. As visualized in \cref{fig:spectral-probing}, we decompose the sequence of values from each embedding dimension (768 in both LMs) using the DCT (\citealp{ahmed1974dct}; DCT-II), weight the appropriate $k$ by a fixed amount or by the learned weight in $\boldsymbol{\gamma}$, before applying the IDCT to reconstruct a sequence of real values. These make up each dimension of the filtered embeddings.

Following \citet{tamkin2020prism}, the original/filtered embeddings are passed to a linear probe \citep{alain2017linear} consisting of two parameters: a transformation $\boldsymbol{W} \in \mathbb{R}^{E\times C}$ and a bias $\boldsymbol{b} \in \mathbb{R}^{C}$, where $E$ is the embedding dimension and $C$ is the number of classes specific to each task.

\paragraph{Training} As we run probing experiments, neither the 108M-parameter BERT, nor the 178M-parameter mBERT are fine-tuned. We only train the linear probe which has 1,538--36,912 parameters depending on the task, plus the 512 parameters of the learned spectral filter $\boldsymbol{\gamma}$. As in \citet{tamkin2020prism}, we use the Adam optimizer \citep{kingma2014adam} with a learning rate of $10^-3$ which decays by 0.5 every time the loss plateaus. Updates are applied in batches of size 32 across a maximum of 30 epochs, with an early stopping patience of 1. Each setup is run with the five random seeds: 1932, 2771, 7308, 8119, 9095. On our hardware consisting of an NVIDIA A100 GPU with 40GBs of VRAM and an AMD Epyc 7662 CPU, training a probe takes approximately 10 minutes.

\paragraph{Evaluation} In order to probe a sequence of contextualized embeddings for information at different timescales, it is necessary to apply each filter at the sub-word level. To measure the effect of different frequencies, we follow ~\newcite{tamkin2020prism} and evaluate all tasks using accuracy (\textsc{Acc}) on the sub-word level. Note that for token-level tasks each token label is therefore repeated across all of its sub-words, while for sequence-level tasks, each sub-word is classified with the label of its sequence.

\paragraph{Implementation} All models are implemented using PyTorch v1.10 \citep{pytorch} and NumPy v1.22 \citep{numpy}. Additionally, we use a modified version of the \texttt{torch-dct} package \citep{ziyang2018torchdct} to perform the DCT and IDCT. Visualizations are generated using matplotlib v3.5 \citep{matplotlib}. Further, the code for reproducing our experiments is available at \href{https://github.com/mainlp/spectral-probing}{https://github.com/mainlp/spectral-probing}.

%
%
\section{Detailed Results}\label{app:detaied-results}

The following supplements the results presented in \cref{sec:experiments} with more detailed scores. \cref{tab:result-details-monolingual} lists the exact scores for the monolingual English experiments on \textsc{PoS} and \textsc{Topic} using the \textsc{Orig} embeddings, the fixed-band filters and the learned \textsc{Auto} filter. \cref{tab:result-details-multilingual} lists the detailed scores for the \textsc{Orig} and \textsc{Auto}-filtered embeddings per language, in addition to the cross-lingual mean and standard deviation, across our seven tasks.

While the scores across random initializations never exceed a standard deviation of 1.0, it is important to note that scores may have higher variance across languages. This is to be expected due to different data across languages as well as pre-training availability. However we note that overall performance patterns (i.e., higher \textsc{Auto} and relative task performance) are consistent across languages.

\begin{table*}
\centering
\begin{tabular}{lrrrrrrr}
\toprule
\textsc{Task} & \multicolumn{1}{l}{\textsc{Orig}} & \multicolumn{1}{l}{\textsc{Low}} & \multicolumn{1}{l}{\textsc{Mid-Low}} & \multicolumn{1}{l}{\textsc{Mid}} & \multicolumn{1}{l}{\textsc{Mid-High}} & \multicolumn{1}{l}{\textsc{High}} & \multicolumn{1}{l}{\textsc{Auto}} \\
\midrule
\textsc{PoS} & 95.8{\footnotesize$\pm$0.1} & 21.9{\footnotesize$\pm$0.0} & 21.8{\footnotesize$\pm$0.1} & 26.2{\footnotesize$\pm$0.1} & 48.6{\footnotesize$\pm$0.1} & 90.6{\footnotesize$\pm$0.0} & 95.9{\footnotesize$\pm$0.0} \\
\textsc{Topic} & 41.3{\footnotesize$\pm$0.2} & 71.2{\footnotesize$\pm$0.4} & 18.4{\footnotesize$\pm$0.3} & 5.6{\footnotesize$\pm$0.3} & 5.6{\footnotesize$\pm$0.3} & 5.6{\footnotesize$\pm$0.4} & 72.1{\footnotesize$\pm$0.3} \\
\bottomrule
\end{tabular}
\caption{\textbf{Detailed Monolingual Results} (\textsc{Acc}) for unfiltered (\textsc{Orig}), low (L), mid-low (ML), mid (M), mid-high (MH), high (H), and automatically learned filters (\textsc{Auto}), on the tasks of \textsc{PoS}-tagging and \textsc{Topic} classification. Reported are the mean over five random initializations $\pm$ standard deviations. The same results plus the spectral profiles (frequency weightings) learned by \textsc{Auto} are plotted in \cref{fig:results-monolingual}.}
\label{tab:result-details-monolingual} 
\end{table*}

\begin{table*}
\centering
\begin{tabular}{ll|rrrrrr|r}
\toprule
\textsc{Task} & \multicolumn{1}{l|}{\textsc{Emb}} & \multicolumn{1}{l}{\textsc{DE}} & \multicolumn{1}{l}{\textsc{EN}} & \multicolumn{1}{l}{\textsc{ES}} & \multicolumn{1}{l}{\textsc{FR}} & \multicolumn{1}{l}{\textsc{JA}} & \multicolumn{1}{l}{\textsc{ZH}} & \multicolumn{1}{|l}{\textsc{Avg}} \\
\midrule
\multirow{2}*{\textsc{PoS}} & \textsc{Orig} & 92.0{\footnotesize$\pm$0.0} & 91.6{\footnotesize$\pm$0.1} & 93.8{\footnotesize$\pm$0.0} & 95.1{\footnotesize$\pm$0.1} & 92.5{\footnotesize$\pm$0.0} & 89.5{\footnotesize$\pm$0.1} & 92.4{\footnotesize$\pm$1.9} \\
 & \textsc{Auto} & 92.1{\footnotesize$\pm$0.1} & 91.6{\footnotesize$\pm$0.0} & 93.9{\footnotesize$\pm$0.0} & 95.1{\footnotesize$\pm$0.0} & 92.7{\footnotesize$\pm$0.1} & 89.8{\footnotesize$\pm$0.1} & 92.5{\footnotesize$\pm$1.8} \\
\midrule
\multirow{2}*{\textsc{Dep}} & \textsc{Orig} & 79.0{\footnotesize$\pm$0.1} & 78.4{\footnotesize$\pm$0.1} & 81.2{\footnotesize$\pm$0.1} & 83.0{\footnotesize$\pm$0.1} & 79.6{\footnotesize$\pm$0.1} & 70.6{\footnotesize$\pm$0.2} & 78.6{\footnotesize$\pm$4.3} \\
 & \textsc{Auto} & 79.5{\footnotesize$\pm$0.2} & 78.4{\footnotesize$\pm$0.1} & 81.8{\footnotesize$\pm$0.1} & 83.8{\footnotesize$\pm$0.1} & 80.8{\footnotesize$\pm$0.1} & 71.3{\footnotesize$\pm$0.2} & 79.3{\footnotesize$\pm$4.3} \\
\midrule
\multirow{2}*{\textsc{NER}} & \textsc{Orig} & 90.3{\footnotesize$\pm$0.0} & 85.3{\footnotesize$\pm$0.1} & 90.4{\footnotesize$\pm$0.0} & 88.1{\footnotesize$\pm$0.1} & 84.1{\footnotesize$\pm$0.1} & 89.5{\footnotesize$\pm$0.0} & 88.0{\footnotesize$\pm$2.7} \\
 & \textsc{Auto} & 90.4{\footnotesize$\pm$0.0} & 85.5{\footnotesize$\pm$0.1} & 90.5{\footnotesize$\pm$0.0} & 88.3{\footnotesize$\pm$0.0} & 84.4{\footnotesize$\pm$0.1} & 89.7{\footnotesize$\pm$0.0} & 88.1{\footnotesize$\pm$2.6} \\
\midrule
\multirow{2}*{\textsc{QA}} & \textsc{Orig} & 63.2{\footnotesize$\pm$0.2} & 64.5{\footnotesize$\pm$0.1} & 64.1{\footnotesize$\pm$0.2} & 63.9{\footnotesize$\pm$0.3} & 61.0{\footnotesize$\pm$0.8} & 60.7{\footnotesize$\pm$0.8} & 62.9{\footnotesize$\pm$1.6} \\
 & \textsc{Auto} & 66.8{\footnotesize$\pm$0.1} & 68.1{\footnotesize$\pm$0.5} & 67.9{\footnotesize$\pm$0.2} & 68.1{\footnotesize$\pm$0.2} & 65.1{\footnotesize$\pm$0.1} & 66.1{\footnotesize$\pm$0.4} & 67.1{\footnotesize$\pm$1.2} \\
 \midrule
\multirow{2}*{\textsc{Senti}} & \textsc{Orig} & 56.0{\footnotesize$\pm$0.2} & 57.1{\footnotesize$\pm$0.2} & 58.7{\footnotesize$\pm$0.2} & 57.1{\footnotesize$\pm$0.2} & 57.2{\footnotesize$\pm$0.2} & 58.0{\footnotesize$\pm$0.2} & 57.4{\footnotesize$\pm$0.9} \\
 & \textsc{Auto} & 64.0{\footnotesize$\pm$0.2} & 63.5{\footnotesize$\pm$0.5} & 65.4{\footnotesize$\pm$0.2} & 64.7{\footnotesize$\pm$0.5} & 65.4{\footnotesize$\pm$0.5} & 62.7{\footnotesize$\pm$0.3} & 64.3{\footnotesize$\pm$1.1} \\
\midrule
\multirow{2}*{\textsc{Topic}} & \textsc{Orig} & 22.7{\footnotesize$\pm$0.1} & 26.8{\footnotesize$\pm$0.4} & 22.9{\footnotesize$\pm$0.3} & 24.0{\footnotesize$\pm$0.3} & 22.9{\footnotesize$\pm$0.5} & 43.3{\footnotesize$\pm$0.4} & 27.1{\footnotesize$\pm$8.1} \\
 & \textsc{Auto} & 34.3{\footnotesize$\pm$0.7} & 39.8{\footnotesize$\pm$0.4} & 30.2{\footnotesize$\pm$0.2} & 30.7{\footnotesize$\pm$0.4} & 35.8{\footnotesize$\pm$0.5} & 52.3{\footnotesize$\pm$0.5} & 37.2{\footnotesize$\pm$8.2} \\
\midrule
\multirow{2}*{\textsc{NLI}} & \textsc{Orig} & 41.5{\footnotesize$\pm$0.2} & 43.6{\footnotesize$\pm$0.3} & 43.2{\footnotesize$\pm$0.2} & 42.7{\footnotesize$\pm$0.2} & 52.3{\footnotesize$\pm$0.3} & 41.7{\footnotesize$\pm$0.2} & 44.1{\footnotesize$\pm$4.1} \\
 & \textsc{Auto} & 51.3{\footnotesize$\pm$0.8} & 56.4{\footnotesize$\pm$0.7} & 54.3{\footnotesize$\pm$0.7} & 54.5{\footnotesize$\pm$0.5} & 67.2{\footnotesize$\pm$0.4} & 53.8{\footnotesize$\pm$1.0} & 56.3{\footnotesize$\pm$5.6} \\
\bottomrule
\end{tabular}
\caption{\textbf{Detailed Multilingual Results} (\textsc{Acc}) for unfiltered (\textsc{Orig}) and automatically learned filters (\textsc{Auto}) on the tasks of \textsc{PoS}-tagging, dependency relation classification (\textsc{Dep}), named entity recognition (\textsc{NER}), question answering (\textsc{QA}), sentiment analysis (\textsc{Senti}), \textsc{Topic} classification, and natural language inference (\textsc{NLI}). Each task covers the languages German (DE), English (EN), Spanish (ES), French (FR), Japanese (JA) and Chinese (ZH). Reported are the mean over five random initializations $\pm$ standard deviations as well as the mean over languages (\textsc{Avg}) $\pm$ the standard deviation across languages. The latter results are reported in \cref{tab:results-multilingual}, in addition to the spectral profiles (frequency weightings) learned by \textsc{Auto} in \cref{fig:filters}.}
\label{tab:result-details-multilingual} 
\end{table*}

\end{document}